# RAC3: Retrieval-Augmented Corner Case Comprehension for Autonomous Driving with Vision-Language Models

Yujin Wang, *Student Member, IEEE*, Quanfeng Liu, Jiaqi Fan, Jinlong Hong, Hongqing Chu, Mengjian Tian*, Bingzhao Gao*, *Member, IEEE* and Hong Chen, *Fellow, IEEE*

***Abstract*—Understanding and addressing corner cases is essential for ensuring the safety and reliability of autonomous driving systems. Vision-language models (VLMs) play a crucial role in enhancing scenario comprehension, yet they face significant challenges, such as hallucination and insufficient real-world grounding, which compromise their performance in critical driving scenarios. In this work, RAC3, a novel framework designed to enhance the performance of VLMs in corner case comprehension, is proposed. RAC3 integrates a frequency-spatial fusion (FSF) image encoder, a cross-modal alignment training method for embedding models with hard and semi-hard negative mining, and a fast querying and retrieval pipeline based on K-Means clustering and hierarchical navigable small world (HNSW) indexing. A multimodal chain-of-thought (CoT) prompting strategy to guide analogical reasoning and reduce hallucinations during inference is introduced. Moreover, an update mechanism is integrated into RAC3 to ensure continual learning within the framework. Extensive experiments on the CODA and nuScenes datasets demonstrate that RAC3 significantly improves corner case comprehension across multiple downstream tasks. Compared to prior state-of-the-art methods, RAC3 achieves the highest final score of 74.46 on the CODA-LM benchmark and shows consistent performance gains when integrated with end-to-end frameworks like DriveLM. These results demonstrate the effectiveness of retrieval-augmented strategies and cross-modal alignment for safer and more interpretable autonomous driving.**

## I. Introduction

### A. Motivation

Despite the significant development achieved in the field of autonomous driving, such systems still lack the ability to comprehend and generalize when facing corner cases, and thus require possible human intervention. The traditional rule-based approach to the development of autonomous driving cannot solve this problem, and lightweight end-to-end neural networks also exhibit significant limitations in scenario comprehension [1], [2]. In recent years, advancements in large-scale machine learning models have propelled the field of embodied intelligence, enabling new paradigms for interaction between artificial systems and their environments. Large models (LMs), ranging from uni-modal architectures focused on textual tasks to sophisticated multimodal systems, have demonstrated exceptional capabilities across a variety of applications. Among these, multimodal large language models (MLLMs), especially vision-language models (VLMs), have emerged as a significant development, leveraging the complementary strengths of textual and visual data to achieve nuanced understanding and reasoning. Such models have been effectively applied to critical areas, including autonomous driving, where precise scenario comprehension, especially corner case comprehension, is paramount for ensuring safety and functionality in complex environments [3], [4], [5], [6], [7], [8], [9].

The integration of LMs into autonomous systems, particularly autonomous driving vehicles, has introduced new challenges and opportunities. These models exhibit robust capabilities in tasks such as object detection, semantic segmentation and trajectory prediction, which are critical for navigating complex urban environments [10], [11], [12], [13], [14], [15], [16]. Moreover, VLMs promise enhanced decision-making through their ability to integrate information from diverse sensory inputs, such as visual streams and textual instructions, creating a more holistic comprehension of the scene [17], [18], [19], [20], [21].

Despite these advancements, VLMs face persistent challenges in corner case comprehension, particularly due to the phenomenon of hallucination. Hallucination refers to instances where models generate outputs that are inconsistent with the real-world content they aim to represent. In the context of autonomous driving, hallucinations can manifest as erroneous object detection, inaccurate attribute descriptions, or implausible relational interpretations within the environment [22]. Such issues not only undermine the reliability of these systems but also pose significant safety risks. For example, a VLM trained on imperfect visual-textual data or visual question answering (VQA) data may erroneously infer the presence of a pedestrian or vehicle that does not exist, leading to potentially dangerous

This research was supported by National Key R&D Program of China (2023YFB2504400), the National Nature Science Foundation of China (No. 62373289 and No. 62473291), Shanghai Municipal Science and Shanghai Automotive Industry Science and Technology Development Foundation (No.2407), Guangdong Basic and Applied Basic Research Foundation (No. 2022A1515110591), Shenzhen Stable Supporting Project (No. 20220718111409001), and the Fundamental Research Funds for the Central Universities. *(Corresponding author: Mengjian Tian (tianmengjian@sztu.edu.cn) and Bingzhao Gao (gaobz@tongji.edu.cn))*

Yujin Wang, Quanfeng Liu, Jinlong Hong, Hongqing Chu and Bingzhao Gao are with the School of Automotive Studies, Tongji University, Shanghai 201804, China.

Jiaqi Fan is with Shanghai Research Institute for Intelligent Autonomous Systems, Tongji University, Shanghai 201210, China.

Mengjian Tian is with the College of Urban Transportation and Logistics, Shenzhen Technology University, Shenzhen 518118, China.

Hong Chen is with the College of Electronic and Information Engineering, Tongji University, Shanghai 201804, China.

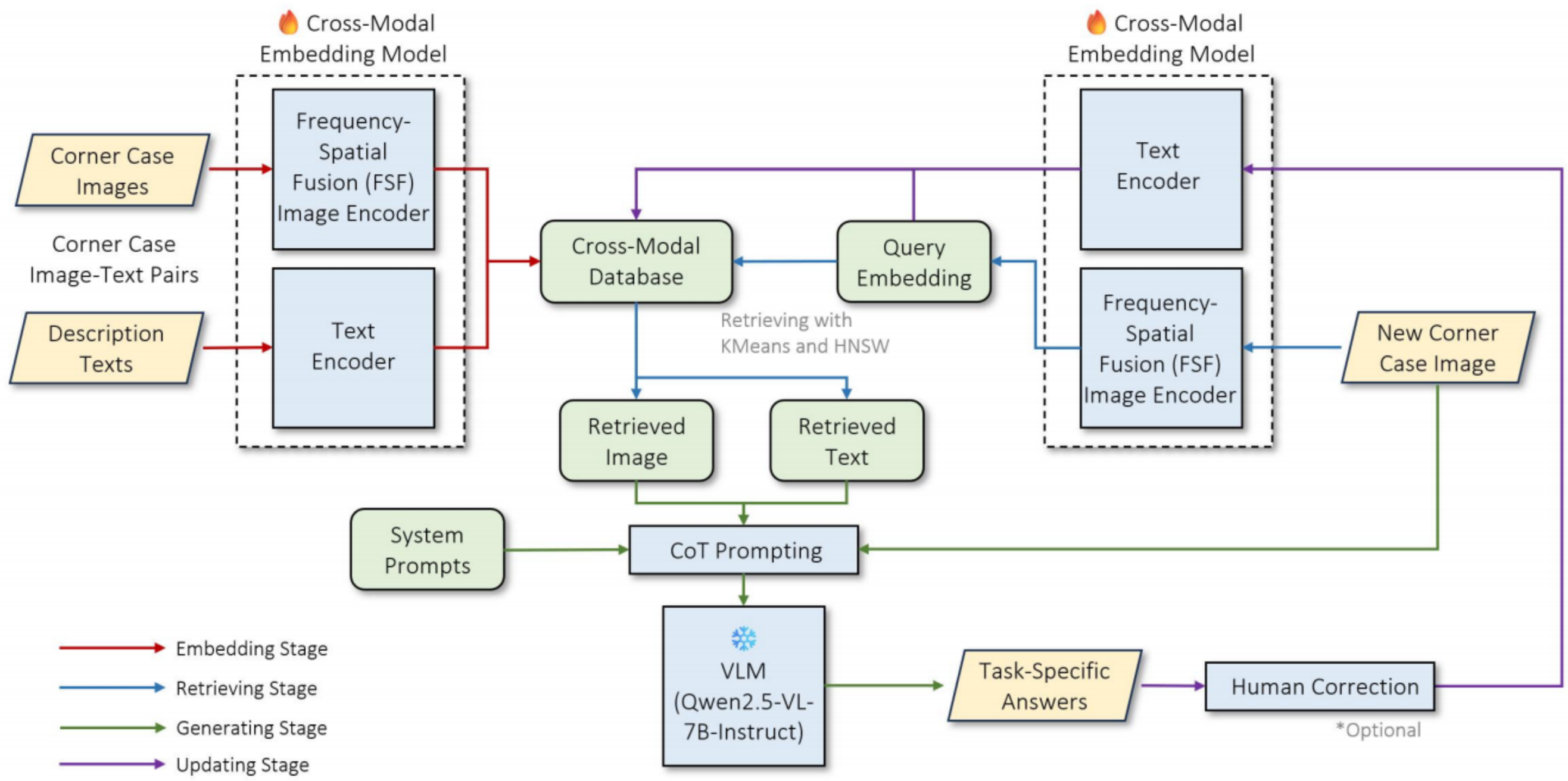

Fig. 1. Overview of the RAC3 framework, which consists of four stages: embedding, retrieving, generating, and updating. The embedding stage encodes corner case image-text pairs into a cross-modal database using a trained cross-modal embedding model. The retrieving stage is triggered by a new corner case image, and the image and text of the most similar corner case are retrieved. Through chain-of-thought (CoT) prompting, the VLM is prompted to generate task-specific answers in the generating stage. The new corner case image and the generated answer can be encoded into the cross-modal database with the updating stage, with human correction optionally introduced when severe hallucinations occur.

decisions [23], [24], [25], [26], [27]. Addressing hallucination is thus critical for ensuring the robustness and applicability of multimodal systems in real-world scenarios [28], [29], [30], [31], [32].

To mitigate hallucination and enhance the fidelity of VLMs, retrieval-augmented generation (RAG) has emerged as a promising framework. RAG combines the generative capabilities of LMs with external retrieval mechanisms to ground model outputs in factual and contextually relevant data [33], [34]. By incorporating real-time retrieval from structured knowledge bases or unstructured datasets, RAG ensures that model predictions are informed by the most relevant evidence, thereby reducing the likelihood of hallucinations. Such capabilities are of particular significance in safety-critical domains like autonomous driving, where decision latency and reliability are paramount.

The application of RAG to scenario comprehension in autonomous driving is especially significant. RAG not only provides a mechanism to validate and refine model predictions but also facilitates the incorporation of domain-specific knowledge, such as traffic regulations and environmental conditions, into the decision-making process [15], [19], [35]. For instance, during navigation in an urban setting, RAG-enabled systems can retrieve contextually relevant data about nearby landmarks, traffic density, or weather conditions to augment the scene comprehension of the model. Furthermore, RAG enhances the interpretability of multimodal models by enabling the traceability of predictions to specific data sources, thereby fostering trust and reliability in autonomous systems. RAG's utility extends beyond hallucination mitigation to support counterfactual reasoning and scenario-based testing in autonomous driving.

It is quite vital that with the enhancement of RAG, the former corner cases that require human takeover could be embedded and added into the existing vector database. The next time when another similar corner case appears, the VLM could get prior knowledge from the database, therefore the human takeover will not be required. This is of great significance for reducing the takeover rate and achieving high-level autonomous driving in the real world.

### B. Contribution

In this work, we propose a retrieval-augmented corner case comprehension (RAC3) framework for autonomous driving, introducing a novel approach to guide VLMs in generating hallucination-mitigated comprehension of rare driving scenarios, as depicted in Figure 1. The main research contributions of this work are outlined as follows:

**A Frequency-Spatial Fusion (FSF) Image Encoder Architecture:** We propose, for the first time within a multimodal RAG architecture, an image encoder that leverages frequency-domain representations extracted via fast Fourier transform (FFT). These representations are systematically integrated with spatial-domain features to enrich the semantic encoding of visual inputs, thereby enhancing the cross-modal retrieval and reasoning capabilities of RAC3 framework.

**A Cross-Modal Alignment Training Method with Hard and Semi-Hard Negative Mining:** To improve the alignment of visual and textual information in the cross-modal embedding model, we employ contrastive learning with hard and semi-hard negative mining. Specifically, the model is trained on the CODA dataset [36] using InfoNCE-based contrastive loss and triplet loss, incorporating semi-hard negative samples during the learning process.

**A Querying and Retrieving Pipeline with K-Means and HNSW Indexing:** To enable fast cross-modal retrieval within a high-dimensional vector space composed of massive data, we introduce a hybrid querying and retrieving pipeline combining K-Means clustering [37] and HNSW indexing [38]. By compressing the sample space via clustering and performing efficient retrieval with HNSW, the cross-modal retrieval latency is reduced to within milliseconds.

**A Chain-of-Thought (CoT) Prompting Strategy:** The core idea of our CoT prompting strategy is to guide multimodal reasoning through analogical comparison, enabling VLMs to transfer structured interpretive patterns from a reference corner case to a new query case, thereby mitigating hallucinations and enhancing comprehension.

**An Update Mechanism for Continual Learning:** In practical deployment, the text generated by the model, together with new corner case images, is incorporated into the original vector database. In cases where the generated outputs exhibit persistent hallucinations, human-in-the-loop corrections are incorporated to ensure the accuracy of the updates. This iterative update mechanism facilitates the continuous enhancement of the system's intelligence during real-world operation.

## II. Related Works

### A. RAG Technologies

The core idea of RAG technology is to introduce an external retrieval module that dynamically retrieves relevant information during the generation process, thereby improving the performance of generative models. In visual-linguistic tasks, RAG effectively compensates for the limitations of scarce knowledge by combining external knowledge bases. The model is not only able to extract information from images but also retrieves supplementary knowledge via the retrieval mechanism, thereby improving the quality and accuracy of the generated output. Jiang et al. [39] propose a RAG-based framework for visual-linguistic models, demonstrating how retrieval-augmented generation significantly enhances model performance in complex tasks, especially those requiring background knowledge. This research indicates that traditional end-to-end VLMs are often limited when faced with insufficient knowledge, whereas RAG, through the incorporation of external knowledge bases, enables the model to integrate more contextual information during the generation process, improving its reasoning and generative abilities.

Building upon this, Shao et al. [40] further explore the application of RAG in VQA tasks. They propose that by integration the retrieval mechanism into pre-trained VLMs, model performance in complex reasoning tasks could be significantly enhanced. Furthermore, Ram et al. [41] study the pre-training and fine-tuning processes of RAG, demonstrating how RAG can further enhance model performance in the fine-tuning stage by incorporating large-scale external data sources during pre-training. RAG not only acquires broader background knowledge during the initial training phase but also effectively utilizes this information during fine-tuning, enhancing the model's cross-modal reasoning ability, especially in cross-modal retrieval tasks, where RAG significantly improves model performance. Meanwhile, Zheng et al. [42] point out that RAG technology not only enhances the model's generative capabilities but also improves its flexibility and adaptability in handling complex multimodal tasks, especially when dealing with tasks lacking sufficient annotations or background knowledge.

As RAG technology continues to deepen its application across various tasks, the key challenge, especially in open-domain VQA tasks, lies in how to dynamically retrieve relevant background knowledge through the retrieval mechanism to improve the model's reasoning accuracy [43]. Although RAG can provide more contextual information, optimizing the retrieval and generation processes, as well as handling the vast amounts of potential external knowledge, remains a current research challenge. To address these issues, Yoran et al. [44] propose a context-aware retrieval-augmented generation method, which adjusts the retrieved content based on the specific needs of the task. This enables the model to more precisely select and utilize external knowledge to tackle complex and variable multimodal tasks.

In multimodal retrieval tasks, Ma et al. [45] further suggest that RAG can serve as a critical mechanism to enhance the processing capabilities of VLMs, particularly in tasks requiring reasoning across multiple modalities. By integrating a retrieval mechanism, RAG can provide more background information for each task, enabling the model to handle a wider range of tasks. Hussien et al. [34] demonstrate how RAG-enhanced VLMs improve cross-modal retrieval, particularly in establishing connections between images and text. RAG effectively utilizes external textual information to significantly improve the accuracy of retrieval.

Regarding performance enhancement, Yuan et al. [33] propose a dynamic knowledge retrieval mechanism, emphasizing that adjusting retrieval and generation processes in real-time, based on task-specific requirements, is crucial for improving RAG model performance. Through this approach, RAG can flexibly select the most relevant background knowledge according to different task demands, thereby achieving better performance in various multimodal tasks. Additionally, Lewis et al. [46] demonstrate the application of RAG in cross-modal retrieval tasks, where, especially in the case of multimodal inputs, dynamic retrieval of relevant information significantly improves the precision and diversity of retrieval results.

### B. Hallucination Mitigation of VLMs

Recent efforts to mitigate hallucinations in VLMs have led to the development of various strategies that target different stages of the model's workflow, including data expansion, model training, and inference correction.

HalluciDoctor [47] introduces a novel cross-checking paradigm to detect semantic hallucinations and generate counterfactual instruction data, thereby enhancing the model's robustness. Similarly, Recaption [48] refines datasets by rewriting captions with the ChatGPT model and fine-tuning the VLMs on these updated datasets, reducing the occurrence of fine-grained hallucinations.

Moreover, several model-training techniques have been explored to reduce hallucinations by improving the model's capabilities in perception and generation. For example, He et al. [49] enhance the VLM by incorporating multiple visual expert models, including object detectors and OCR, to enrich the model's knowledge base. Jain et al [50] further improve the model's object perception by providing additional visual inputs such as segmentation and depth maps. Chen et al. [51] introduces a model that injects spatially aware and semantically rich visual evidence into the VLM, enhancing its multimodal understanding. Moreover, Jiang et al. [52] apply contrastive learning, treating hallucinated texts as hard negative samples to better align visual and textual representations.

In addition to these data and training-based methods, post-hoc corrections during the inference stage also play a critical role in alleviating hallucinations [53]. For example, VCD [54] employs a visual contrastive strategy during decoding, comparing output distributions from both original and distorted visual inputs to ensure consistency between the generated content and the visual data. LogicCheckGPT [55] creates a logical closed-loop method using object-to-attribute and attribute-to-object inquiring to verify consistency, while Volcano [56] takes an iterative approach to reduce multimodal hallucinations, applying a critique-revision-decision cycle during the inference.

## III. Method: RAC3

### A. Frequency-Spatial Fusion Image Encoding Model

To more effectively extract key features from dynamic scenarios in autonomous driving, we propose a joint frequency-spatial image encoding model, as illustrated in Figure 2. First, fast Fourier transform (FFT) is applied to the input image to extract frequency-domain features, such as road boundaries and key information related to specific traffic participants. The input image $I$ is transformed into the frequency domain using FFT as defined in (1):

$$\mathcal{F}(I)(u,v) = \sum_{x=0}^{H-1} \sum_{y=0}^{W-1} I(x,y)\, e^{-\,i\,2\pi\left(\frac{ux}{H} + \frac{vy}{W}\right)}, \qquad (1)$$

where $I(x,y) \in \mathbb{R}^{H\times W\times 3}$ is the RGB input image of height $H$ px and width $W$ px, $\mathcal{F}(I)(u,v)$ is the complex-valued frequency spectrum at frequency coordinates $(u,v)$, and $i$ is the imaginary unit.

The amplitude spectrum $A = |\mathcal{F}(I)|$ is primarily extracted, preserving the energy distributions of both high- and low-frequency information, and the phase spectrum $P = \angle\mathcal{F}(I)$. While both amplitude and phase spectra are derived from FFT, we opt to utilize only the amplitude spectrum in our encoder, since the amplitude spectrum captures the energy distribution across frequencies, which is closely associated with perceptual saliency and structural patterns (e.g., lane boundaries or object edges) in driving scenes. Moreover, the phase spectrum, although essential for reconstructing spatial information, is highly sensitive to local pixel variations and often lacks stability across visually similar scenarios, making it less robust for learning cross-modal embeddings. Therefore, in our framework, only the amplitude spectrum is used for frequency-domain attention, ensuring both efficiency and robustness in feature fusion.

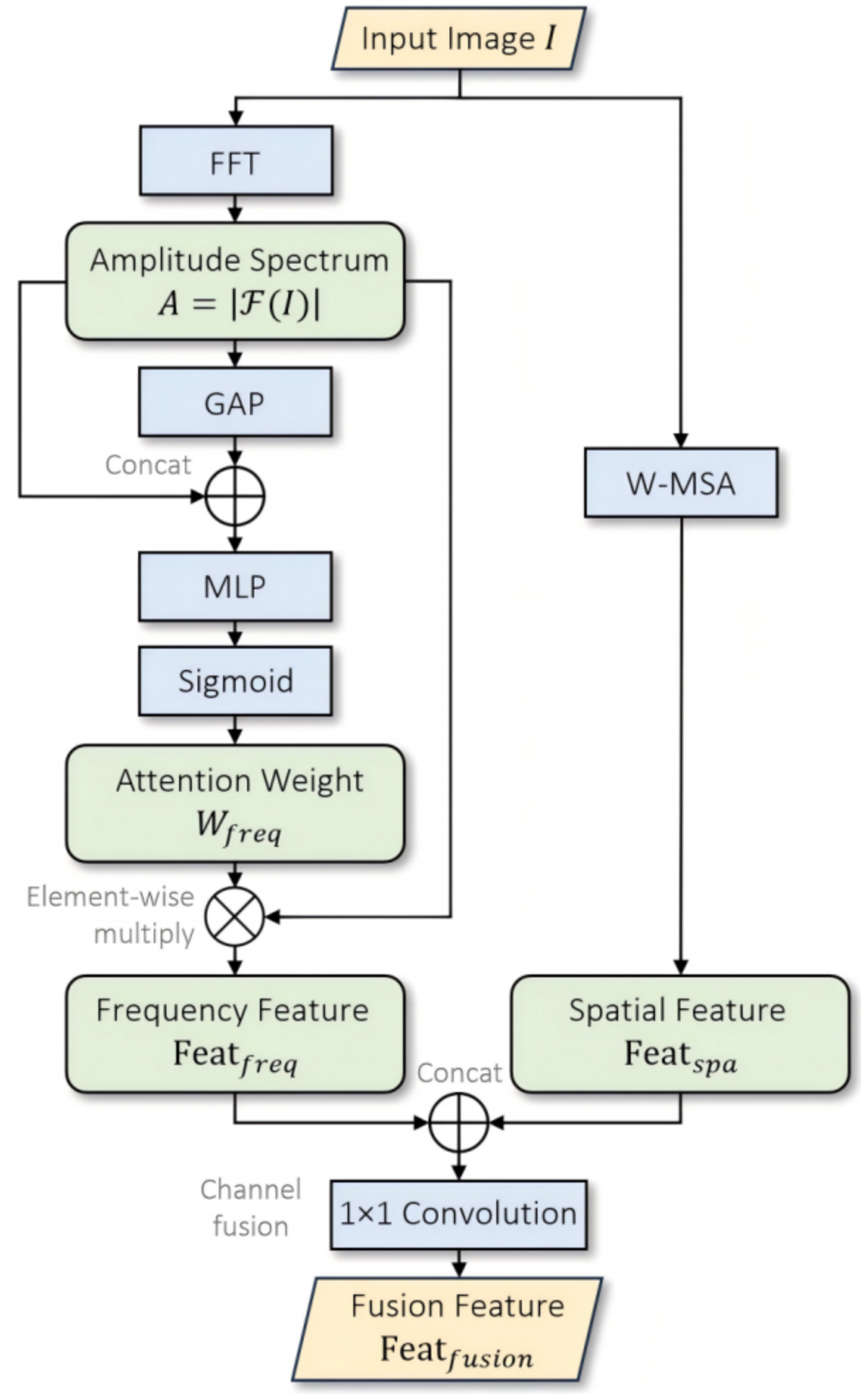

Fig. 2. Architecture of the proposed frequency-spatial fusion (FSF) image encoder, which integrates frequency- and spatial-domain features for enhanced visual representation.

In the practical implementation, the input image is partitioned into fixed-size blocks (e.g., $16\times16$). Each block is then processed individually via the FFT to obtain its corresponding local amplitude spectrum.

Furthermore, to enable the network to automatically attend to more salient frequency-domain features, we design an attention module based on global average pooling (GAP) and multi-layer perceptrons (MLPs):

$$W_{\text{freq}} = \sigma\big(\text{MLP}\big(A \oplus \text{GAP}(A)\big)\big), \qquad (2)$$

where:

- $\oplus$ denotes concatenation along the channel dimension;
- GAP globally compresses the information in the amplitude spectrum;
- $\sigma$ is Sigmoid activation function, which is used to normalize the attention weights.

A three-layer MLP (for instance, with a hidden dimensionality of 512) is designed. After applying global average pooling, the pooled features is concatenated with the original features to generate the frequency-domain attention weights. The frequency-domain features are derived via element-wise multiplication:

$$\mathrm{Feat_{freq}} = W_{\mathrm{freq}} \otimes A. \tag{3}$$

By contrast, the method employed to extract spatial-domain features is relatively more conventional. The window-based multi-head self-attention (W-MSA) mechanism is leveraged in the Swin Transformer [57] to extract spatial features from the input images:

$$\mathrm{Attention}(Q, K, V) = \mathrm{Softmax}\Big(\frac{QK^{\top}}{\sqrt{d_k}}\Big)\, V. \tag{4}$$

where $d_k$ is the channel dimension of each self-attention head ($d_k = C/h$; for the Swin-Tiny setting we use $C = 96$ and $h = 3$, hence $d_k = \mathbf{32}$).

In this mechanism, $Q$, $K$, and $V$ represent the query, key, and value matrices, respectively, while $d_k$ denotes the feature dimension of each head. We adopt the Swin Transformer (Tiny configuration, e.g., 4 layers with a window size of 7×7) to implement window-based multi-head self-attention. Through this local self-attention mechanism, fine-grained spatial information is captured from the image, thereby yielding spatial-domain features:

$$\mathrm{Feat_{spa}} = \mathrm{W\text{-}MSA}(I). \tag{5}$$

Finally, features from the frequency and spatial domains are fused. A 1×1 convolution is then applied for channel integration:

$$\mathrm{Feat_{fusion}} = \mathrm{Conv}_{1\times1}\big(\mathrm{Feat_{freq}} \oplus \mathrm{Feat_{spa}}\big). \tag{6}$$

This frequency-spatial fusion (FSF) model preserves the global responses inherent in the frequency-domain information while also accommodating the local characteristics of spatial details, thereby yielding a fused feature representation for subsequent cross-modal alignment training.

### *B. Cross-Modal Embedding Model Alignment Training Framework*

Cross-modal alignment is crucial in the vector embedding stage. Poor alignment may lead to a decline in model performance and weakened generalization ability [58]. In this work, a novel cross-modal embedding model alignment training framework is introduced, which aligns visual and textual embeddings within a shared latent space. The architecture comprises two primary encoding branches-one for images and one for text-followed by a uniform projection stage and contrastive alignment objectives.

**Image encoder.** As introduced in the last subsection, the frequency-spatial fusion (FSF) image encoding model that processes each image $\mathbf{X}$ into feature maps is employed:

$$\mathbf{V} = \mathrm{Feat_{fusion}}(\mathbf{X}). \tag{7}$$

An adaptive average pooling operation reduces each feature map to a 64-dimensional vector, which is then projected into a $d$-dimensional embedding space via a fully connected layer. The resulting embedding is subsequently $l_2$-normalized to obtain:

$$\mathbf{e}_i^{\mathrm{img}} = \frac{\mathbf{W}_{\mathrm{img}}\tilde{\mathbf{v}}_i + \mathbf{b}_{\mathrm{img}}}{\|\mathbf{W}_{\mathrm{img}}\tilde{\mathbf{v}}_i + \mathbf{b}_{\mathrm{img}}\|_2}, \tag{8}$$

where $\tilde{\mathbf{v}}_i \in \mathbb{R}^{64}$ is the pooled image feature and $\mathbf{W}_{\mathrm{img}} \in \mathbb{R}^{d\times 64}$, $\mathbf{b}_{\mathrm{img}} \in \mathbb{R}^{d}$ are learnable parameters, with $d = 256$ in all experiments.

**Text encoder.** For the text branch, we adopt the bge-base-en-v1.5 model [59] for text embedding, since it achieves high accuracy on NLP benchmarks through its multi-task learning framework while ensuring efficient feature extraction and low computational cost, while other embedding models typically incur higher resource consumption or lack comparable multi-task performance.

The text encoder produces a 768-dimensional embedding for each input text $s_i$. Each text embedding is projected into the same $d$-dimensional space by a fully connected layer, followed by $l_2$-normalization:

$$\mathbf{e}_i^{\mathrm{txt}} = \frac{\mathbf{W}_{\mathrm{txt}}\mathbf{t}_i + \mathbf{b}_{\mathrm{txt}}}{\|\mathbf{W}_{\mathrm{txt}}\mathbf{t}_i + \mathbf{b}_{\mathrm{txt}}\|_2}, \tag{9}$$

where $\mathbf{t}_i \in \mathbb{R}^{768}$ is the raw output, and $\mathbf{W}_{\mathrm{txt}}$, $\mathbf{b}_{\mathrm{txt}}$ are trainable parameters of the projection layer.

**Contrastive loss.** We first adopt a symmetrical variant of the InfoNCE-based contrastive loss [60], originally popularized for learning representations in contrastive self-supervised frameworks. Let $\mathbf{e}_i^{\mathrm{img}}, \mathbf{e}_j^{\mathrm{txt}} \in \mathbb{R}^d$ denote the normalized image and text embeddings for the $i$-th and $j$-th samples in a batch, respectively, with $d$ being the common embedding dimension. A logit matrix is constructed:

$$\mathbf{Z} \in \mathbb{R}^{B\times B}, \quad Z_{ij} = \frac{\mathbf{e}_i^{\mathrm{img}} \cdot \mathbf{e}_j^{\mathrm{txt}}}{\tau}, \tag{10}$$

where $\mathbf{e}_i^{\mathrm{img}}, \mathbf{e}_j^{\mathrm{txt}} \in \mathbb{R}^d$ are the L2-normalized embeddings for the $i$-th image and $j$-th text sample, respectively, and $\tau$ is a temperature hyperparameter (**0.05** unless otherwise stated) and $B$ is the batch size (fixed to **32** in all experiments). Treating the diagonal entries $Z_{ii}$ as positive matches ($i$-th image paired with $i$-th text) and off-diagonal entries as negative matches, the contrastive loss comprises two parts, namely the Image-to-Text Alignment $\mathcal{L}_{\mathrm{i2t}}$ and the Text-to-Image Alignment $\mathcal{L}_{\mathrm{t2i}}$:

$$\mathcal{L}_{\mathrm{i2t}} = -\sum_{i=1}^{B} \log\left(\frac{\exp(Z_{ii})}{\sum_{j=1}^{B}\exp(Z_{ij})}\right), \tag{11}$$

$$\mathcal{L}_{\mathrm{t2i}} = -\sum_{i=1}^{B} \log\left(\frac{\exp(Z_{ii})}{\sum_{j=1}^{B}\exp(Z_{ji})}\right). \tag{12}$$

The combined contrastive loss is then the average of these two terms:

$$\mathcal{L}_{\mathrm{con}} = \frac{1}{2}(\mathcal{L}_{\mathrm{i2t}} + \mathcal{L}_{\mathrm{t2i}}). \tag{13}$$

By enforcing alignment from both the image and text perspectives, this formulation captures a symmetrical matching criterion, penalizing inconsistent cross-modal pairs more effectively.

**Hard negative triplet loss.** While the InfoNCE-based contrastive loss effectively separates positives and negatives, it does so with respect to all in-batch negatives. To further emphasize especially hard negatives, a margin-based triplet loss is incorporated. Using the same logit matrix $\mathbf{Z}$ as above, we let the diagonal entries $Z_{ii}$ represent the similarity of matched pairs, and the off-diagonal entries $Z_{ij} (j \neq i)$ represent the similarity of mismatched pairs. We focus on the hardest negative for each sample by selecting:

$$\max_{j \neq i} Z_{ij} \quad \text{and} \quad \max_{j \neq i} Z_{ji}, \tag{14}$$

i.e., the negative pairs that yield the largest similarity scores, posing the greatest risk of confusion with the positive pair. Therefore, the triplet loss can be written as:

$$\begin{aligned} \mathcal{L}_{\text{triplet}} = \frac{1}{B} \sum_{i=1}^{B} \Big[ & \max(0, \alpha + \max_{j \neq i} Z_{ij} - Z_{ii}) \\ & + \max(0, \alpha + \max_{j \neq i} Z_{ji} - Z_{ii}) \Big], \end{aligned} \tag{15}$$

where $\alpha$ is a margin hyperparameter. By penalizing positive-negative pairs whose similarity is nearly as high as the positive-positive pair, the hard negative triplet loss encourages the learned embeddings to maintain a stricter separation between matching and non-matching samples.

The overall loss could be defined as the combination of the InfoNCE-based contrastive loss and the triplet loss:

$$\mathcal{L} = \mathcal{L}_{\text{con}} + \lambda \mathcal{L}_{\text{triplet}}, \tag{16}$$

where $\lambda$ scales the influence of hard negative triplet learning. In our training process, $\alpha$ is set to 0.3 and $\lambda$ is set to 0.1.

**Semi-hard negative mining.** In addition to leveraging the hardest negatives, the framework optionally supports a semi-hard negative mining strategy, which focuses on moderately challenging negative samples. This approach attempts to strike a balance between easy negatives, which may not provide much training signal, and extremely hard negatives, which may lead to unstable gradients early in training.

Concretely, a similarity matrix $\mathbf{S} \in \mathbb{R}^{B \times B}$ is computed, analogous to $\mathbf{Z}$ but optionally without temperature scaling. For each sample $i$, the code identifies the first negative sample $j \neq i$ such that

$$S_{ij} < 1 - \alpha, \tag{17}$$

where $\alpha$ is the same margin used in the triplet loss. If no such sample is found, the procedure falls back to the least similar entry in the row. During the early stage of training, when the model is not yet well-initialized, semi-hard mining can simply select a random negative to avoid overfitting on outliers. Once these semi-hard negatives are identified, the system applies the regular contrastive loss $\mathcal{L}_{\text{con}}$ solely on the chosen semi-hard negative pairs, thus devoting computational resources to learning from more informative training examples. Note that while triplet loss is applied over the hardest negatives, the semi-hard negative mining strategy is used exclusively to construct training pairs for the contrastive loss $\mathcal{L}_{\text{con}}$, rather than for triplet-based learning.

**Overall training strategy.** The proposed training algorithm supports the following three strategies:

- Employing only the InfoNCE-based contrastive loss.
- Utilizing a combination of contrastive loss and a hard negative triplet loss.
- Applying semi-hard negative mining followed by computing the contrastive loss.

### C. Cross-modal Embedding, Retrieving and Generating

**Step 1: Embedding.** The embedding pipeline implements a cross-modal framework designed to align corner case images and text descriptions within a shared latent space. The unified cross-modal model, which is trained with contrastive loss and triplet loss, projects both image and text embeddings into a common 256-dimensional space.

By processing paired image-text inputs through this architecture, the model generates corresponding embeddings that can be used for downstream retrieval tasks. The image embeddings serve as keys mapping to their associated text embeddings.

For instance, the training set of CODA-LM is embedded, which contains 4,884 corner case images and corresponding descriptions with this embedding pipeline. The size of the cross-modal database is only 45.1 MB. Also, with a larger dataset of the region perception task in CODA-LM, the size of the database is 100.2 MB with 20,414 image-text pairs. The size of the database is related to the length of the text. Therefore, the size of the database is not positively correlated with the number of image-text pairs.

**Step 2: Querying and retrieving.** To facilitate efficient retrieval on large-scale datasets, directly computing the cosine similarity for every embedding is computationally expensive. In some prior RAG pipelines [61], [62], simple similarity-based search is applied directly over the entire embedding space. However, this approach often suffers from poor scalability in large-scale autonomous driving datasets. To ensure fast and accurate retrieval in large-scale multimodal databases, we design a hybrid retrieval pipeline based on K-Means [37] clustering and HNSW (Hierarchical Navigable Small World) [38] indexing. This design is motivated by two key challenges in cross-modal retrieval for autonomous driving:

- The need to quickly localize relevant reference cases from a high-dimensional embedding space under real-time constraints.
- The desire to scale retrieval to large corner case repositories without sacrificing accuracy.

K-Means clustering enables coarse-grained partitioning to narrow down the search space, while HNSW offers efficient approximate nearest neighbor (ANN) search with high recall and sublinear query time. This combination allows RAC3 to maintain both high retrieval fidelity and low latency, which are crucial for autonomous driving.

Given a set of normalized embeddings $\{\mathbf{z}_i\}_{i=1}^{N}$ extracted using a cross-modal encoding module, the pipeline proceeds as follows:

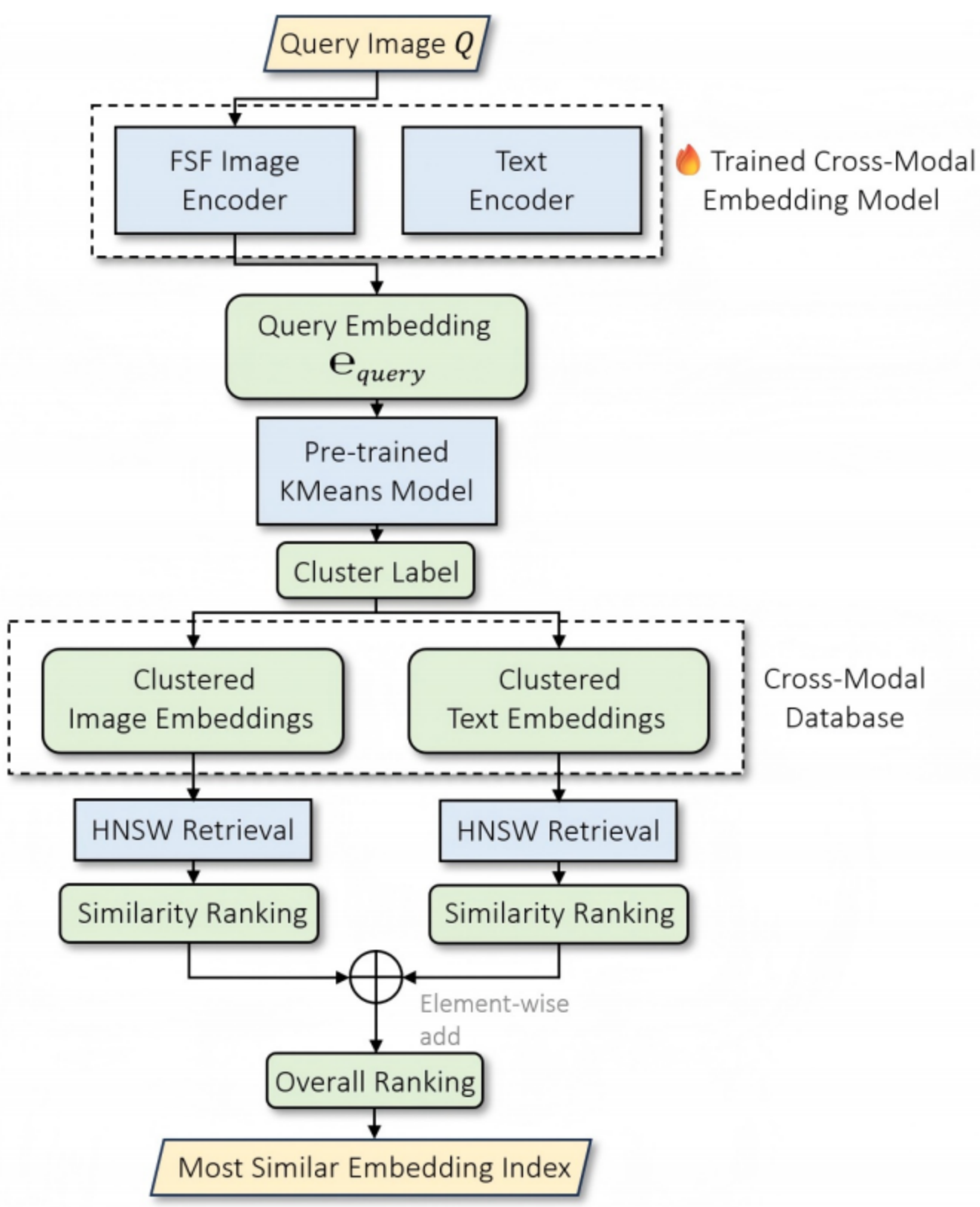

Fig. 3. The querying and retrieval pipeline employed in the RAC3 framework.

*Clustering via K-Means:* The database embeddings are partitioned into $K$ clusters (e.g., $K = 10$) by solving the following optimization problem:

$$\min_{\{\mu_k\}_{k=1}^{K}} \sum_{k=1}^{K} \sum_{\mathbf{z}_i \in C_k} \|\mathbf{z}_i - \mu_k\|^2, \tag{18}$$

where $\mu_k$ is the centroid of cluster $C_k$. This step groups similar embeddings together, effectively reducing the search space for subsequent nearest neighbor queries.

*HNSW index construction:* For each cluster, an HNSW index is constructed using the embeddings belonging to that cluster. The index is configured to use cosine distance, defined as:

$$d(\mathbf{q}, \mathbf{z}_i) = 1 - \mathbf{q}^\top \mathbf{z}_i, \tag{19}$$

where $\mathbf{q}$ is a query vector. The performance and accuracy of the HNSW index are controlled by several key parameters:

- $M$: The maximum number of connections (neighbors) allowed per node in the graph. A higher $M$ typically creates a more densely connected graph, improving search accuracy at the expense of increased memory consumption and longer construction times. $M$ is set to 16, since it is a commonly adopted compromise that ensures high retrieval accuracy without imposing excessive resource demands.
- $ef_construction$: This parameter determines the trade-off between index construction time and the quality of the graph. A larger $ef_construction$ usually results in a more accurate index, though it requires additional time during the construction phase. $ef_construction$ is set to 200, since it is a widely used and effective value that helps ensure robust retrieval performance in subsequent query phases.
- $ef$: Used during the query phase, this parameter controls the search depth. Increasing $ef$ often improves recall (i.e., the likelihood of finding the true nearest neighbor) but may lead to slower query speeds. $ef$ is set to 50, since it is often found to provide a favorable balance between query speed and retrieval accuracy.

In our cross-modal database, K-Means clustering and HNSW index construction are performed independently for image and text embeddings.

*Query processing:* As illustrated in Figure 3, when a query image is provided, its embedding is computed using the cross-modal model. The embedding of the query image is used to perform HNSW retrieval separately within the image and text sections of the cross-modal database. The pre-trained K-Means model predicts the cluster label for this query embedding.

The corresponding HNSW index for that cluster is then used to perform a nearest neighbor search, returning the similarity ranking indices of all image and text embeddings in the database. The rankings are then summed, and the image-text embedding pair with the highest combined ranking is selected as the final retrieval result, returning the global index of this embedding ($top - k = 1$). In the ablation studies, we discuss the difference of model performance with various $top - k$ values.

*Result retrieval:* The retrieved global index is used to directly obtain the associated text description from the origin text file, yielding the final retrieval result. This approach leverages the balanced clustering of K-Means to build compact HNSW indices, ensuring that nearest neighbor searches are performed over smaller, more relevant subsets of the data. The careful tuning of HNSW parameters ($M$, $ef_construction$, and $ef$) allows for a flexible balance between retrieval accuracy and efficiency, rendering the method highly suitable for real-time, cross-modal retrieval tasks in large-scale autonomous driving datasets.

**Step 3: Generating.** In this work, a multimodal chain-of-thought (CoT) prompting strategy is applied to guide the VLM in understanding complex corner cases from the perspective of the ego vehicle and therefore in generating specific answers. Unlike conventional text-only CoT prompting, since the reference image and text have already been retrieved, our method incorporates both visual and textual exemplars to facilitate few-shot generalization in real-world perception tasks.

As shown in Figure 4, the prompting scheme is designed to mimic human-like analogical transfer: the VLM is first exposed to the retrieved corner case comprising its image and task-specific description. This pairing acts as a demonstration example, showing the expected reasoning structure and decision-relevant elements.

Subsequently, the query image is presented and the VLM is prompted to generate a structured interpretation based on the implicit analogy to the reference. According to different down-stream tasks, the VLM must identify various requirements and the difference between the query and reference images. Generally, the VLM should evaluate the key objects' appearance, spatial location, orientation, and behavior relevance.

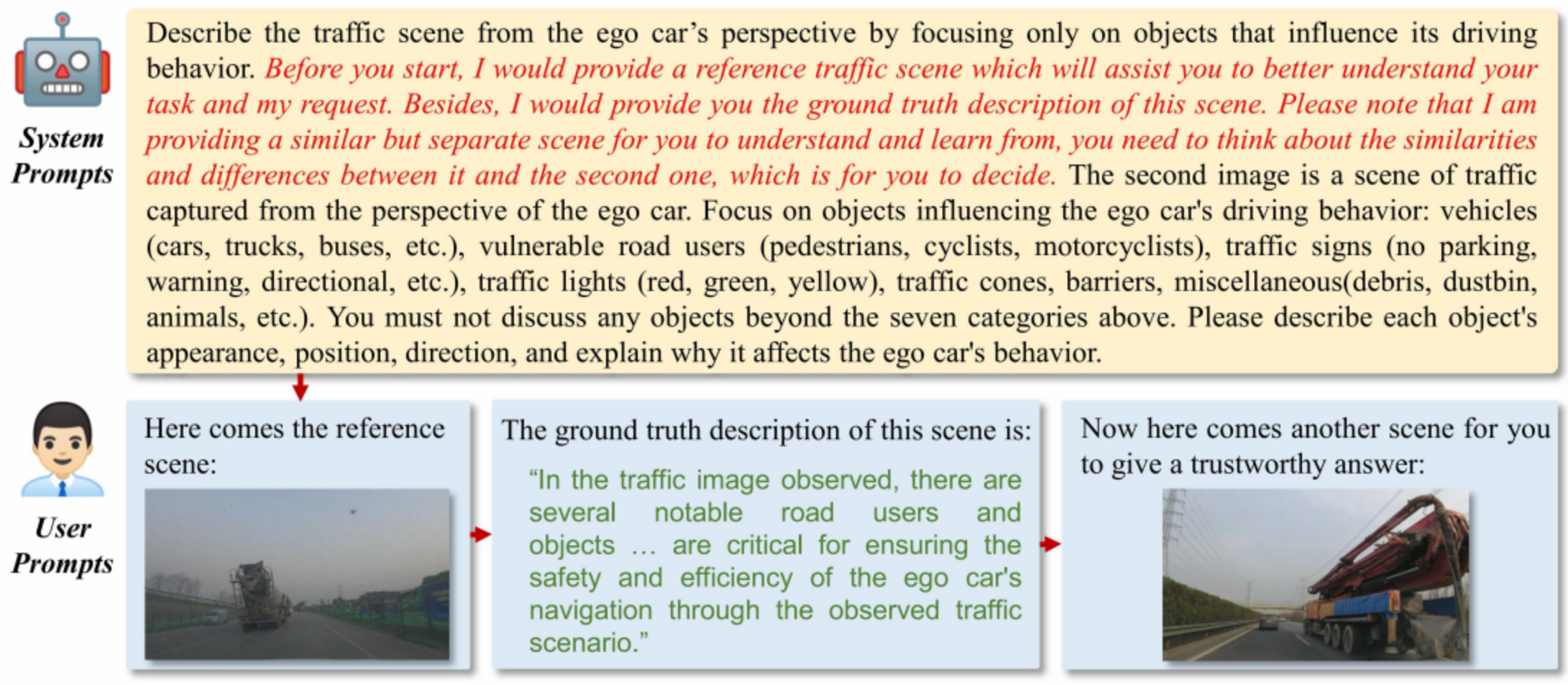

Fig. 4. Multimodal chain-of-thought (CoT) prompting strategy used in the RAC3 method. It encourages the VLM to draw implicit comparisons between the reference and query corner cases step by step, enabling more trustworthy responses.

This few-shot multimodal CoT prompting encourages step-by-step, interpretable output generation, enabling the VLM to transfer reasoning patterns from known to new corner cases through analogical inference. Our experiments, which will be discussed in Section IV, show this design significantly enhances reasoning quality and semantic precision in open-domain corner case comprehension tasks.

Furthermore, a memory and update mechanism is incorporated. The responses generated through CoT prompting are jointly encoded with the query image and subsequently embedded into the original database as new key-value pairs. To further enhance the reliability, an optional human-in-the-loop correction interface is integrated into the pipeline to mitigate the risk of severe hallucinations that may still arise in the responses generated by the VLM. In the absence of manual intervention, the generated responses are directly encoded and integrated into the existing database. This constitutes a real-time post-processing procedure. To ensure the objectivity and fidelity of our results, no manual corrections are applied to the generated responses during subsequent quantitative experiments. Instead, some authentic corner cases are presented to demonstrate the efficacy of the proposed update mechanism.

## IV. Experiments

### A. Implementation

As mentioned before, we implement the cross-modal embedding model using PyTorch and utilize Swin Transformer [57] and bge-base-en-v1.5 [59] as our backbones. The AdamW optimizer [63] is utilized with distinct learning rates assigned to different components of the model:

- The image encoder's parameters are assigned a learning rate of 1e-3, which facilitates rapid learning on the image side.
- The projection layers for both the image and text modalities use a learning rate of 3e-4.
- The text encoder is given a smaller learning rate of 1e-5 to prevent excessive updates to the pre-trained model.

Besides, a global weight decay of 0.01 is applied to regularize the model parameters. The total numbers of training epochs are set to 50, 75, 100, 150, 200, 300, respectively. A batch size of 32 is used for data loading, and the data used for training is the original validation split of CODA-LM dataset [36], [64], which contains 4,384 corner case images and corresponding GPT-generated and human-revised ground truth descriptions. Training our cross-modal embedding model to convergence takes approximately 13 hours (200 epochs) on a single NVIDIA RTX 3090 GPU.

### B. Evaluation

**Evaluation datasets.** Two mainstream datasets, CODA [36] and nuScenes [74], are primarily applied to evaluate the performance of our RAC3 method in handling corner case scenarios. The CODA dataset comprises approximately 10,000 meticulously selected real-world driving scenes and focuses on object-level corner cases. The nuScenes dataset is a large-scale, multimodal dataset for autonomous driving and comprises 1,000 driving scenes collected from urban environments in Boston and Singapore. In both these two datasets, well-annotated textual descriptions are provided. These two datasets are selected according to each task's requirement.

**Evaluation benchmark.** According to the adopted datasets, the CODA-LM [64] benchmark is utilized, which is a well-established benchmark for corner case comprehension in autonomous driving.

CODA-LM is constructed on CODA dataset and contains high-quality human-annotated textual information for three downstream tasks: *general perception*, *region perception* and *driving suggestion*. The general perception task aims to demonstrate the comprehensive understanding of critical road key entities in corner cases. The region perception task measures VLMs' capabilities to understand corner case objects with the bounding boxes provided. The driving suggestion task is used to evaluate the capability of VLMs in generating driving advice.

TABLE I
QUANTITATIVE COMPARISON OF DIFFERENT METRICS AMONG VARIOUS SOTA BASELINES AND OUR RAC3 METHOD ON THE CODA-LM BENCHMARK

| Method | Base Model | Techniques | General Perception | Region Perception | Driving Suggestion | Final Score |
|---|---|---|---|---|---|---|
| - | LLaVA-v1.5-7B [65] | - | 19.30 | 42.06 | 23.16 | 28.17 |
| - | InternVL-V1-5-20B [66] | - | 38.38 | 61.53 | 41.18 | 47.03 |
| - | Qwen2.5-VL-7B-Instruct [67] | - | 44.54 | 51.27 | 47.82 | 47.88 |
| - | GPT-4V [7] | - | 57.50 | 56.26 | 63.30 | 59.02 |
| RAD [68] | Qwen2.5-VL-7B-Instruct [67] | Fine-tuning | 47.16 | 67.54 | 57.25 | 57.32 |
| CODA-VLM [36] | LLaVA1.5 (on Llama-3-8B) [65] | Fine-tuning | 55.04 | 77.68 | 58.14 | 63.62 |
| NexusAD [61] | InternVL2-26B [66] | Fine-tuning & RAG | 57.58 | 84.31 | 65.02 | 68.97 |
| OpenDriver [62] | MiniCPM-8B [69] | Fine-tuning & RAG | 54.41 | 83.00 | 71.76 | 69.72 |
| FNN [70] | GPT-4 [7] | CoT Prompting | <u>59.00</u> | <u>84.37</u> | 70.80 | 71.39 |
| llmforad [71] | LLaVA-NeXT-7B [72] & GPT-4 [7] | Fine-tuning (LLaVA) | 58.70 | 83.41 | <u>74.26</u> | <u>72.12</u> |
| **RAC3 (ours)** | **Qwen2.5-VL-7B-Instruct [67]** | **RAG & CoT Prompting** | **64.29** | **84.48** | **74.60** | **74.46** |

TABLE II
QUANTITATIVE DEMONSTRATION OF PERFORMANCE ENHANCEMENT OF DRIVELM MODELS WITH RAC3 ON THE CODA-LM BENCHMARK

| DriveLM Model [73] | RAC3 Integration | General Perception | Region Perception | Driving Suggestion | Final Score |
|---|---|---|---|---|---|
| DriveLM (on LLaMA-LoRA-BIAS-7B) | × | 45.76 | 69.28 | 48.92 | 54.65 |
| DriveLM (on LLaMA-LoRA-BIAS-7B) | ✓ | **54.53** | **73.24** | **57.73** | **61.83** |
| DriveLM (on LLaMA-BIAS-7B) | × | 47.42 | 71.67 | 52.38 | 57.16 |
| DriveLM (on LLaMA-BIAS-7B) | ✓ | **56.29** | **75.25** | **59.06** | **63.53** |
| DriveLM (on LLaMA-CAPTION-7B) | × | 50.43 | 74.57 | 55.48 | 60.16 |
| DriveLM (on LLaMA-CAPTION-7B) | ✓ | **57.84** | **77.48** | **61.41** | **65.58** |

CODA-LM consists of 9,768 corner case images, which are divided into a training set (4,884 images), a validation set (4,384 images) and a test set (500 images). The original validation set is used as our training set for cross-modal embedding model as mentioned before. We embed the original training set with 4,884 images and their corresponding annotated textual descriptions into vector databases, and evaluate the performance of our method on the original test set. The evaluation process aligns with the original settings of CODA-LM. Given reference ground truths of the test set and few-shot In-Context-Learning samples, GPT-4 is used as the judge to evaluate the correctness of VLMs' response with a score ranging from 1 to 10. The score for each task is calculated as the arithmetic mean of the scores obtained by running all images in the test set with VLMs. Potential hallucination of GPT-4 is revised by human annotators. The final score is calculated using the following formula:

$$\begin{aligned}\text{Final-Score}_{codalm} = \frac{1}{3}\Big(&\text{GPT-Score}_{\text{General Perception}} \\ &+ \text{GPT-Score}_{\text{Region Perception}} \\ &+ \text{GPT-Score}_{\text{Driving Suggestions}}\Big). \quad (20)\end{aligned}$$

In our evaluation, all scores are standardized to a 1-to-100 scale to facilitate readability and comparison. All inference processes are conducted on a server with four A800 GPUs.

**Comparison with other methods.** We compare RAC3 to 10 baselines, namely 4 vanilla VLMs and 5 SOTA methods on the CODA-LM benchmark on CODA dataset, as listed in Table I. Among the SOTA methods, CODA-VLM [36], NexusAD [61], OpenDriver [62] and llmforad [71] involve fine-tuning of the base model, and most of these methods adopt LoRA [75], which is a common method for fine-tuning VLMs. Some methods, like NexusAD and OpenDriver, also integrate RAG technology, but the retrieval principles are rather simple in their solutions. In NexusAD, a pre-trained InternViT-6B serves as the image encoder without task-specific training, and the speed of retrieval is not mentioned. OpenDriver adopts a RAG paradigm by leveraging visual embeddings from segmented image regions to retrieve semantic information from a vision-language database. Retrieved textual descriptions are further used to refine captions and support automatic question-answer pair generation, enhancing the multimodal dataset's richness and reasoning capability. Their RAG approach emphasizes data generation for fine-tuning rather than directly guiding the model to generate answers. As for FNN, this method solely relies on CoT prompting to guide GPT-4 in generating answers. It cannot be deployed locally and is heavily dependent on network connectivity.

Our method is based on the Qwen2.5-VL-7B-Instruct

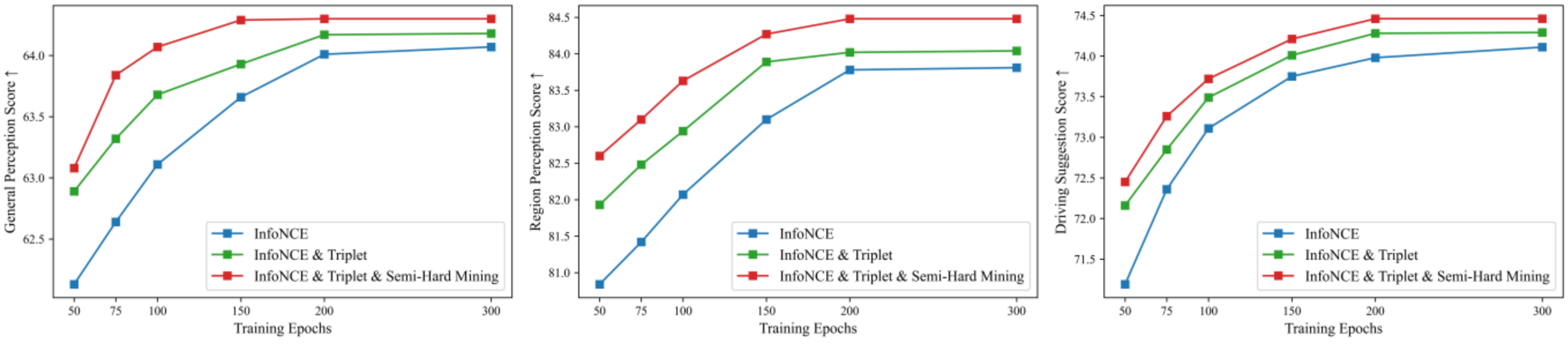

Fig. 5. Ablation studies on various training strategies and epochs of cross-modal embedding model alignment training. We adopt the CODA-LM [64] benchmark as the metrics.

model, which has a relatively small number of parameters and can be deployed locally on consumer-grade GPUs. On the CODA-LM benchmark, RAC3 outperforms all baseline methods, achieving a 55.51% final score improvement compared to the vanilla Qwen2.5-VL-7B-Instruct, demonstrating its effectiveness convincingly.

To further contextualize the improvements brought by RAC3, we compare it with RAD [68], a fine-tuning-based method using the same base model, Qwen2.5-VL-7B-Instruct. RAD relies on fine-tuning to improve the model's spatial perception performance, while RAC3 adopts a plug-in retrieval-augmented approach that enhances reasoning without modifying the base model parameters. This makes RAC3 more lightweight and easier to deploy. Moreover, RAC3 demonstrates better overall performance across diverse tasks by leveraging structured retrieval and analogical reasoning, highlighting its superior generalization and interpretability over fine-tuning-based methods.

**Further exploration of the integration of RAC3 into E2E autonomous driving.** To further demonstrate the potential of the RAC3 framework within end-to-end autonomous driving systems, we conduct experiments on the nuScenes dataset by integrating RAC3 into the DriveLM framework. Therefore, we conduct experiments on CODA-LM benchmark primarily to demonstrate that our framework can significantly enhance the performance of DriveLM algorithms in end-to-end autonomous driving, as shown in Table II, highlighting its potential for integration into other established methods. We complement the ground truth descriptions with ChatGPT-4o [7] of all scenes in nuScenes dataset with further human revision.

The results reveal that RAC3 integration consistently improves performance across all four evaluation metrics regardless of the underlying DriveLM model. Notably, the model DriveLM (on LLaMA-CAPTION-7B) with RAC3 integration achieves the highest Final Score of 65.58, compared to 60.16 without RAC3. Similarly, the model DriveLM (on LLaMA-BIAS-7B) sees an increase from 57.16 to 63.53, indicating a robust and generalizable improvement across different base models.

It is worth noting that DriveLM remains frozen throughout all experiments, and RAC3 operates as a plug-in module without any fine-tuning of the base VLM. This design underscores RAC3's compatibility and low-cost integration capability for enhancing existing vision-language models in autonomous driving.

These consistent gains suggest that RAC3's cross-modal embedding module plays a critical role in enhancing situational awareness and decision-making capabilities, which are essential for E2E autonomous driving. The improvements in Region Perception and Driving Suggestion, in particular areas closely related to corner case comprehension and action planning, underscore RAC3's ability to improve multimodal grounding in dynamic driving environments.

This performance uplift on the CODA-LM benchmark strongly supports the feasibility of integrating RAC3 into real-world E2E autonomous driving frameworks such as DriveLM. Given that the RAC3 framework can operate efficiently with relatively lightweight models like Qwen2.5-VL-7B-Instruct, it shows high potential for deployment on consumer-grade hardware, which is crucial for practical applications.

## C. Ablation Studies

In ablation studies, we mainly discuss the influence of various training strategies of the cross-modal embedding model, and various k values of top-k retrievals. The CODA-LM benchmark is adopted in this subsection, and Qwen2.5-VL-7B-Instruct also serves as the base model. All inference processes are conducted on a server with four A800 GPUs. In addition, we also validate the performance improvement of our combined K-Means & HNSW retrieval strategy in comparison with simple similarity-based search and vanilla HNSW on our constructed dataset.

**Ablation studies on training strategies.** As illustrated in Figure 5, ablation studies on training strategies provide a comprehensive evaluation of how different loss formulations of cross-modal embedding model alignment training affect the VLM's performance on the CODA-LM benchmark.

Three configurations are compared: training with only the InfoNCE-based contrastive loss, combining contrastive loss with hard negative triplet loss, and incorporating semi-hard negative mining on top of the previous setup. The results show a consistent and progressive improvement across all three strategies. RAC3 using the cross-modal embedding model trained solely with InfoNCE-based contrastive loss achieves the lowest performance, suggesting that relying on basic contrastive alignment is insufficient for complex multimodal reasoning tasks such as corner case comprehension. Introducing the hard negative triplet loss leads to a significant performance boost, indicating the effectiveness of explicitly penalizing negative samples to enhance embedding discrimination.

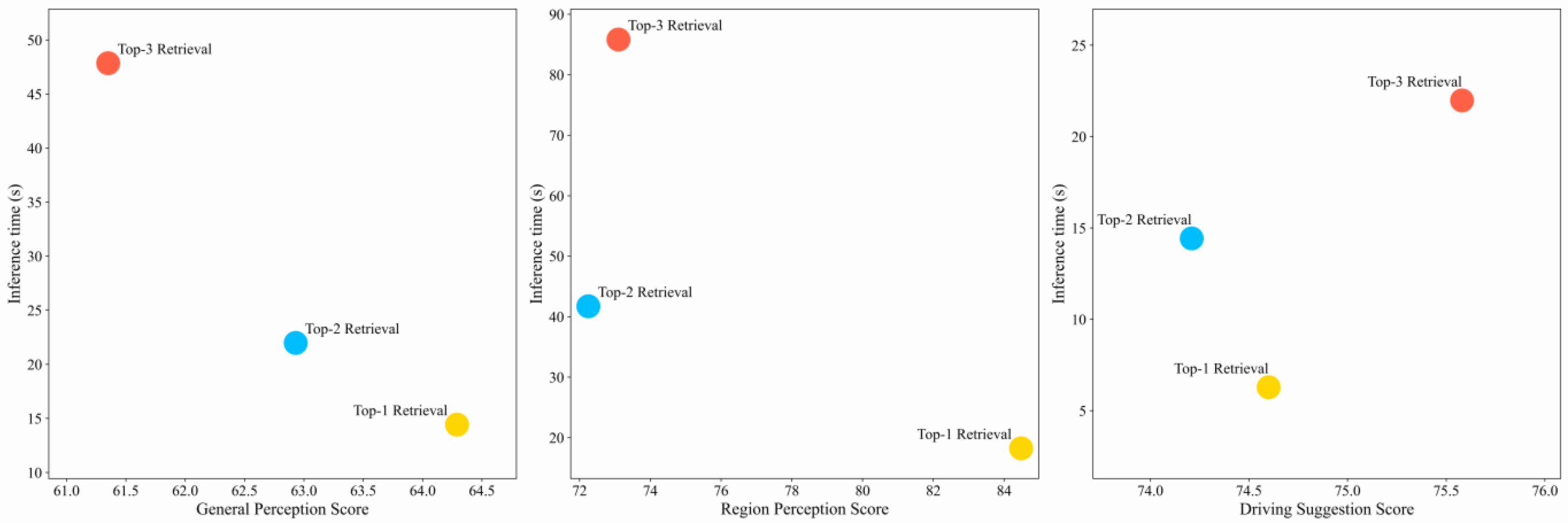

Fig. 6. Ablation studies on various top-k selections. We conduct experiments on k values from 1 to 3 due to restrictions on computational resources. We adopt the CODA-LM [64] benchmark as the metrics.

Building further on this, the inclusion of semi-hard negative mining yields the best overall performance in all metrics. This strategy balances the learning signal by avoiding both trivially easy and excessively hard negative samples, thereby promoting more stable and effective training dynamics. The training process typically converges within 200 epochs.

**Ablation studies on various top-k selections.** Fig. 6 presents the ablation studies on the effect of varying the top-k retrieval parameter in the retrieval pipeline of RAC3 framework. Specifically, the experiment evaluates the model's performance when retrieving the top-1, top-2, and top-3 most similar corner case from the cross-modal database during inference on the CODA-LM benchmark.

In the General Perception and Region Perception tasks, model performance tends to degrade as the value of k increases. This may be attributed to the limited capacity of the relatively small-scale model, which struggles to process an excessive amount of reference information in these tasks. In the Driving Suggestion task, the model achieves its best performance when using top-3 retrieval.

However, for VLMs, inference time is a critical factor that cannot be overlooked. Across all tasks, inference time increases significantly with larger top-k values. This is due to the inclusion of more reference texts and images, which leads to a greater number of input tokens and, consequently, longer processing time for the model. Specifically, in the General Perception task, inference time rises from 14.43 seconds at top-1 to 47.86 seconds at top-3; in the Region Perception task, it increases from 18.24 seconds to 85.84 seconds; and in the Driving Suggestion task, it grows from 6.28 seconds to 21.98 seconds. Although increasing top-k can improve performance in certain tasks, the additional inference time and substantial computational overhead often outweighs the benefits.

Larger top-k values result in significantly higher GPU memory consumption during inference, which exceeds our computational capacity. Moreover, inference time continues to increase with larger k. Due to these limitations, we do not conduct experiments under such settings. Nevertheless, the ablation results suggest that under the RAC3 framework, employing top-1 retrieval is sufficient to achieve optimal performance.

**Ablation studies on various querying and retrieval pipelines.** To validate the effectiveness of our retrieval strategy, we compare the proposed K-Means & HNSW pipeline with simple similarity-based search and vanilla HNSW. As detailed in Table III, while all approaches achieve similar retrieval quality (within $\pm 0.4$ Final Score in the downstream generating stage on the CODA-LM benchmark), our method reduces average retrieval time per query from 200 milliseconds to 7.2 milliseconds, resulting in an approximate 27.8$\times$ speedup in comparison with simple similarity-based search. Additionally, we evaluate vanilla HNSW without clustering, which achieves slightly slower retrieval at 11 milliseconds per query. All retrieval operations are executed on a single CPU (13th-generation Intel® Core™ i7-13700) with no GPU acceleration.

TABLE III
PERFORMANCE IMPROVEMENT OF K-MEANS & HNSW RETRIEVAL PIPELINE

| Retrieval Strategy | Final Score | Average Retrieval Time |
|---|---|---|
| Simple similarity-based search | 74.09 | 200ms |
| Vanilla HNSW | 74.35 | 11ms |
| **K-Means & HNSW** | **74.46** | **7.2ms** |

The improvement is critical for real-time scenario comprehension in autonomous driving. The results confirm that our hybrid approach offers a favorable enhancement in retrieval speed.

We also experiment with modifying certain parameters in the retrieval process, such as the number of clusters and the parameters of HNSW. However, due to the relatively small size of the database, these changes do not lead to significant variations in retrieval time, which remains consistently between 0.007 and 0.008 seconds.

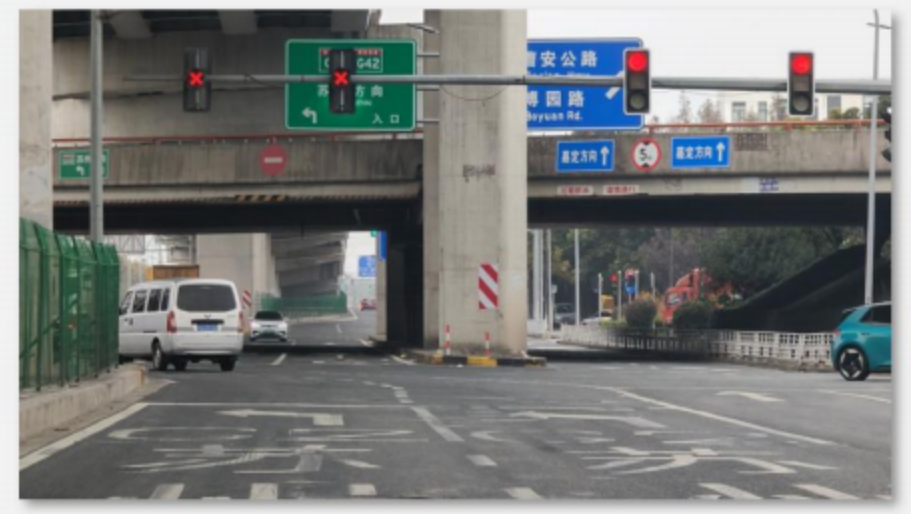

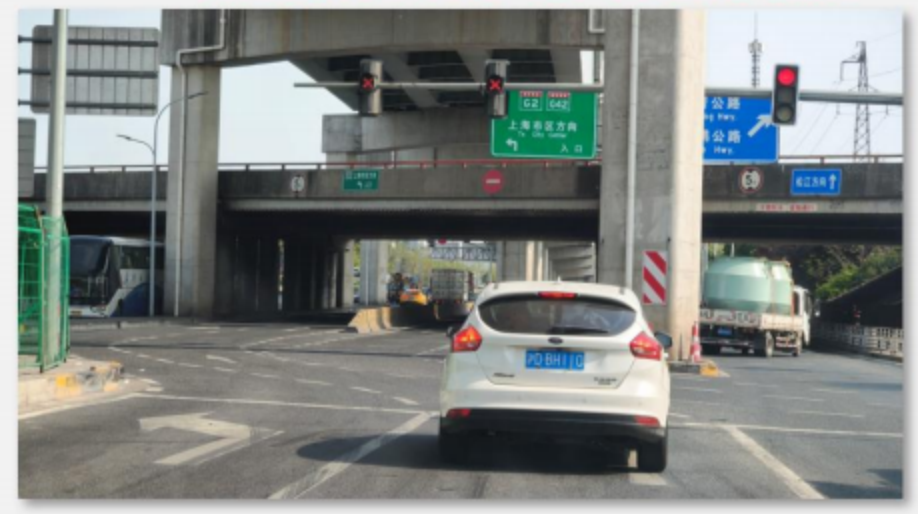

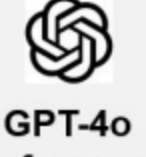

Fig. 7. A representative demonstration of a corner case. Firstly, when an unseen corner case is encountered, the VLM generates descriptions with hallucination, and thus human correction is required. Next, this corner case and the manual correction of the description are embedded and added into the database. Finally, when a similar corner case is encountered, the VLM is able to generate descriptions that closely align with the reference texts produced by GPT-4o, demonstrating effective hallucination mitigation.

### D. Representative Corner Case Demonstration

In the previous experiments, it has been proved that using a fixed corner case database and RAG technology can enhance the scenario comprehension ability of VLMs and mitigate hallucination. In this demonstration, we aim to demonstrate that RAG technology can also facilitate the integration of new corner case information. Through human intervention and correction of descriptions, when encountering this corner case or similar traffic scenarios for the second time, VLMs can obtain prior information through RAG, thereby mitigating hallucination.

As shown in Figure 7, a corner case on urban roads in Shanghai, China is selected, where there is a special-shaped traffic light with a red cross in the middle of the road. It has been observed that most human drivers passing by here for the first time are unable to correctly understand its meaning and thus choose to stop and wait at the intersection. In fact, this traffic light resembles those typically seen at tunnel entrances or highway toll stations, which means that you cannot continue to go straight in this lane but should drive towards the left front, just as indicated by the white arrows painted on the ground.

Since there is no similar driving scenario in our initial database, even with RAG, the Qwen2.5-VL-7B-Instruct VLM also exhibits a similar hallucination and misinterprets it as a signal to stop. Given the known meaning of this signal, the corresponding description can be manually corrected. The corrected description and the image of the new corner case are added to the existing RAG database for future reference.

Subsequently, another corner case with this kind of special-shaped traffic light is taken as the input and invokes the entire system. It is observed that the VLM successfully corrected the previous hallucination and provided the correct interpretation of the traffic light. In addition, the generated descriptions are well-aligned with the reference texts generated by GPT-4o. We test the performance of this update mechanism on 10 pairs of real-world corner cases on urban roads in Shanghai, and this mechanism successfully corrected 8 out of 10 previous hallucinations in similar cases.

Our method exhibits robust performance in both zero-shot and few-shot scenarios. This indicates that when encountering new corner cases, the use of RAG enables the VLM to obtain external knowledge references dynamically instead of integrating the new data with the original pre-trained data and retraining the VLM, thus mitigating hallucination and enhancing its generalization ability.

## V. Conclusion

This paper presents RAC3, a retrieval-augmented framework designed to enhance corner case comprehension in autonomous driving using vision-language models. By introducing a frequency-spatial fusion encoder, contrastive alignment with hard and semi-hard negative mining, an efficient K-Means & HNSW retrieval pipeline, and a CoT-prompting strategy,

RAC3 effectively mitigates hallucinations and improves multimodal grounding.

Extensive experiments on CODA-LM benchmark, CODA and nuScenes datasets demonstrate that RAC3 consistently outperforms existing baselines across multiple tasks, achieving a new state-of-the-art on CODA-LM. Notably, RAC3 operates as a plug-in enhancement module and does not require fine-tuning of the base vision-language model. When integrated into end-to-end systems such as DriveLM, which remains frozen throughout all experiments, RAC3 still delivers significant performance gains, highlighting its practical viability for deployment in real-world autonomous driving systems.

Future research will explore extending RAC3 to broader perception and planning tasks, with the aim of further improving the safety and interpretability of autonomous driving systems through retrieval-augmented strategies.

## BIOGRAPHIES

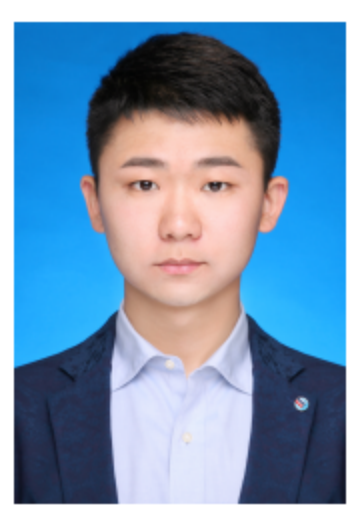

**Yujin Wang** (larswang@tongji.edu.cn) is currently a PhD student at School of Automotive Studies, Tongji University, Shanghai 201804, China. He received his B.S. degree in vehicle engineering from Tongji University, Shanghai, China, in 2023. His research interests include autonomous driving, vision-language models, retrieval-augmented generation and deep learning. He is a student member of IEEE.

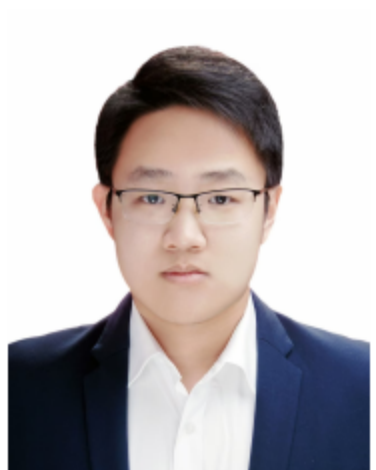

**Quanfeng Liu** (2053657@tongji.edu.cn) is currently working toward his B.S. degree at School of Automotive Studies, Tongji University, Shanghai 201804, China. His research interests include autonomous driving and foundation models.

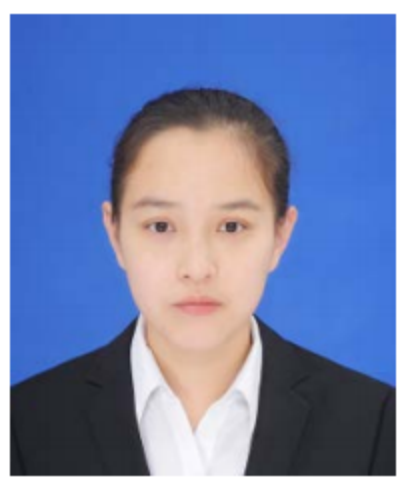

**Jiaqi Fan** (fanjq@tongji.edu.cn) received the B.S. degree from Jilin University, Changchun, China, in 2019, and the M.S. degree from Jilin University, Changchun, China, in 2022. She is currently a Ph.D. student at Shanghai Research Institute for Intelligent Autonomous Systems in Tongji University, China. Her main research interests include scene understanding of autonomous vehicles, vision-language model, and deep learning.

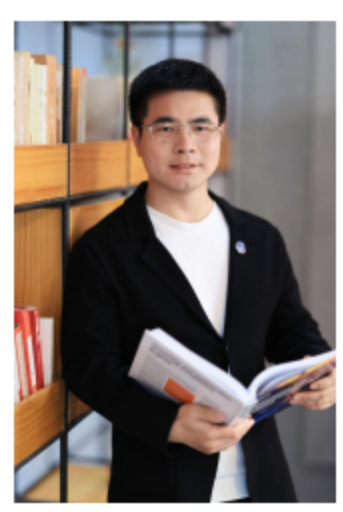

**Jinlong Hong** (hongjl@tongji.edu.cn) received the B.S. degree in vehicle engineering from Yanshan University, Hebei, China, in 2014, and the Ph.D. degree in vehicle engineering from Jilin University, Changchun, Jilin, China, in 2019. He is currently a Post-Doctoral Fellow with the School of Automotive Studies, Tongji University, Shanghai, China. His research interest is mainly on the autonomous driving embodied intelligence.

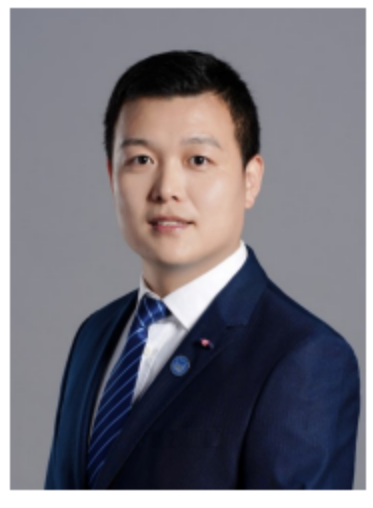

**Hongqing Chu** (chuhongqing@tongji.edu.cn) received the B.S. degree in Vehicle Engineering from Shandong University of Technology, China, in 2012 and his Ph.D. degree in Control Theory and Control Engineering from Jilin University, China, in 2017. He is currently an Associate Professor with the School of Automotive Studies, Tongji University. His current research interests include autonomous driving, vehicle motion control, and advanced driver-assistance system.

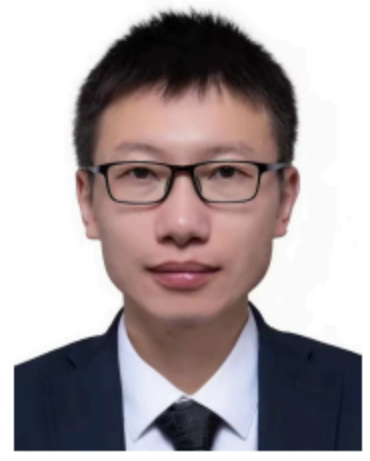

**Mengjian Tian** (tianmengjian@sztu.edu.cn) received the Ph.D. degree in vehicle engineering from Jilin University, Changchun, China, in 2021. He is currently an Assistant Professor with the College of Urban Transportation and Logistics, Shenzhen Technology University, Shenzhen, China. His research interest is in intelligent vehicle chassis mechanisms, dynamics, and control.

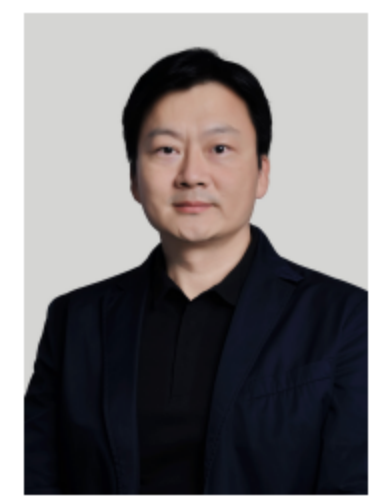

**Bingzhao Gao** (gaobz@tongji.edu.cn) received the B.S. and M.S. degrees in vehicle engineering from Jilin University, Changchun, China, in 1998 and 2002, respectively, and the Ph.D. degree in mechanical engineering from Yokohama National University, Yokohama, Japan, and in control engineering from Jilin University, in 2009. He is currently a Professor with the School of Automotive Studies, Tongji University, Shanghai, China. His research interests include vehicle powertrain control and vehicle stability control. He is a Member of IEEE.

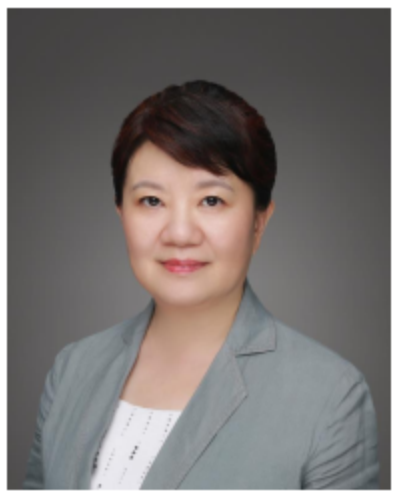

**Hong Chen** (chenhong2019@tongji.edu.cn) received the B.S. and M.S. degrees in process control from Zhejiang University, Hangzhou, China, in 1983 and 1986, respectively, and the Ph.D. degree in system dynamics and control engineering from the University of Stuttgart, Stuttgart, Germany, in 1997. In 1986, she joined the Jilin University of Technology, Changchun, China. From 1993 to 1997, she was a Wissenschaftlicher Mitarbeiter with the Institut fuer Systemdynamik und Regelungstechnik, University of Stuttgart. Since 1999, she has been a Professor with Jilin University, Changchun, and hereafter a Tang Aoqing Professor. From 2015 to 2019, she served as the Director of the State Key Laboratory of Automotive Simulation and Control, Jilin University. She currently joins Tongji University, Shanghai, China, as a Distinguished Professor. Her current research interests include model predictive control, nonlinear control, and applications in mechatronic systems focusing on automotive systems. She is a Fellow of IEEE.